\documentclass{article}

\usepackage{microtype}
\usepackage{graphicx}
\usepackage{subcaption}
\usepackage{booktabs} % for professional tables

\usepackage{hyperref}

\usepackage{placeins}
\usepackage[preprint]{icml2026}
\usepackage{multirow}
\usepackage{multicol}

\newcommand{\perf}[2]{#1{\scriptsize±#2}}
\newcommand{\ul}[1]{\underline{#1}}

\newif\ifshowcomments
\showcommentstrue
\ifshowcomments
\newcommand{\mynote}[2]{\textcolor{blue}{\fbox{\bfseries\sffamily\scriptsize#1}}
  \textcolor{blue}{{$/*$\textsf{\emph{#2}}$*/$}}}
\ifshowcomments

\else
\newcommand{\mynote}[2]{}
\fi
\usepackage{amsmath}
\usepackage{amssymb}
\usepackage{mathtools}
\usepackage{amsthm}

\usepackage[capitalize,noabbrev]{cleveref}

\theoremstyle{plain}

\theoremstyle{definition}

\theoremstyle{remark}

\usepackage[textsize=tiny]{todonotes}

\icmltitlerunning{The Patient is not a Moving Document:
A World Model Training Paradigm for Longitudinal EHR}

\begin{document}

\twocolumn[
  \icmltitle{The Patient is not a Moving Document: \\
A World Model Training Paradigm for Longitudinal EHR}

  \icmlsetsymbol{equal}{*}
  
  \begin{icmlauthorlist}
    \icmlauthor{Irsyad Adam}{equal,comp}
    \icmlauthor{Zekai Chen}{equal,comp}
    \icmlauthor{David Laprade}{comp}
    \icmlauthor{Shaun Porwal}{comp}
    \icmlauthor{David Laub}{comp}
    \icmlauthor{Erik Reinertsen}{comp}
    \icmlauthor{Arda Pekis}{comp}
    \icmlauthor{Kevin Brown}{comp}
  \end{icmlauthorlist}

  \icmlaffiliation{comp}{Standard Model Biomedicine}

  \icmlcorrespondingauthor{Irsyad Adam}{irsyad@standardmodel.bio}
  \icmlcorrespondingauthor{Zekai Chen}{zach@standardmodel.bio}

  \icmlkeywords{Machine Learning, ICML}

  \vskip 0.3in
]

% Add the equal contribution footnote after \maketitle
\begingroup
\renewcommand\thefootnote{}\footnote{*Equal contribution}
\addtocounter{footnote}{-1}
\endgroup


\printAffiliationsAndNotice{}  % no special notice (required even if empty)

\begin{abstract}
Large language models (LLMs) trained with next-word-prediction have achieved success as clinical foundation models. Representations from these language backbones yield strong linear probe performance across  biomedical tasks, suggesting that patient semantics emerge from next-token prediction at scale. However, this paradigm treats patients as a document to be summarized rather than a dynamical system to be simulated; a patient's trajectory emerges from their state evolving under interventions and time, requiring models that simulate dynamics rather than predict tokens. To address this, we introduce SMB-Structure, a world model for structured EHR that grounds a joint-embedding prediction architecture (JEPA) with next-token prediction (SFT). SFT grounds our model to reconstruct future patient states in token space, while JEPA predicts those futures in latent space from the initial patient representation alone, forcing trajectory dynamics to be encoded before the next state is observed. We validate across two large-scale cohorts: Memorial Sloan Kettering (23,319 oncology patients; 323,000+ patient-years) and INSPECT (19,402 pulmonary embolism patients). Using a linear probe evaluated at multiple points along the disease trajectory, we demonstrate that our training paradigm learns embeddings that capture disease dynamics not recoverable by autoregressive baselines, enabling SMB-Structure to achieve competitive performance on complex tasks characterized by high patient heterogeneity. Model weights are available at \texttt{standardmodelbio/SMB-v1-8B-Structure}.
\end{abstract}

\section{Introduction}

The established paradigm for clinical foundation models inherits a core assumption that predicting the next token is a sufficient objective for learning predictive, generalizable representations. Currently, this assumption has proven to be effective, as clinical language models trained on domain-specific medical knowledge achieve strong transfer to downstream prediction tasks \citep{steinberg2021language, wornow2023ehrshot}, pass medical licensing examinations \citep{singhal2023large}, and generate coherent radiology reports from images \citep{chen2024advancing, li2024llavamed}. The empirical success suggests that patient-relevant semantics are recoverable from representations optimized purely for next-word-prediction.

\begin{figure*}[!ht]
\begin{center}
\includegraphics[width=0.75\textwidth]{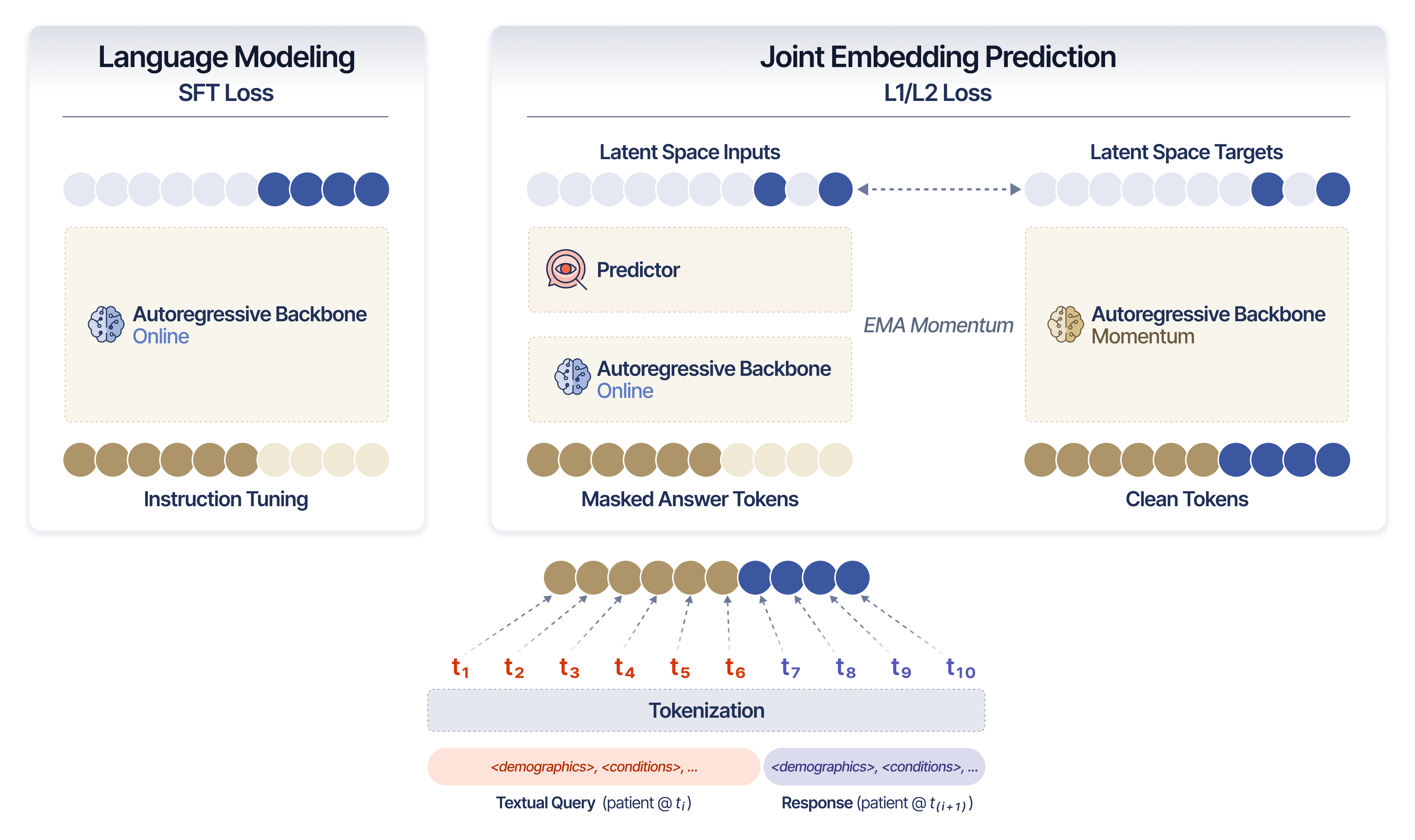}
\end{center}
\caption{Architecture for SMB-Structure for Time-to-Event EHR Modeling.}
\label{fig:arch}  
\end{figure*}

However, the core ideology of reconstruction as opposed to simulation in terms of modeling disease trajectory are fundamentally different. Autoregressive language models trained to predict the next token in clinical text are not designed to understand how a patient's disease manifestation will evolve; they instead model the distributional regularities in clinical documentation~\citep{willig2022can, zevcevic2023causal}. This starkly contrasts with real-world clinical decision-making, where a domain expert evaluates a patient record to anticipate the future course of disease~\citep{prosperi2020causal, sanchez2022causal}. In such settings, the central question is not “what word comes next?” but rather “given the current clinical context, how will the patient’s state evolve?”  Addressing this requires explicit modeling of dynamics and trajectory evolution under observed treatments, rather than static sequence prediction. The training paradigm of autoregressive models enable sufficient learning to answer the first question, but nothing in the training objective incentivizes them to answer the second \citep{kiciman2023causal}. The representation may capture what the patient \emph{is} without encoding where they are \emph{going} \citep{richens2020improving}. 

This gap between reconstruction and dynamics has been recognized in other domains. In robotics, world models learn to predict how latent states evolve under actions, enabling planning without task-specific supervision \citep{ha2018world, hafner2023mastering}. In video understanding, Joint-Embedding Predictive Architectures (JEPA) predict masked content in representation space rather than pixel space, learning features that encode motion and physical plausibility rather than texture \citep{bardes2024revisiting, assran2025vjepa2}. This principle has since been extended to multimodal and language domains: VL-JEPA demonstrates that predicting aligned vision–language embeddings enables grounded cross-modal reasoning without explicit generative supervision \citep{bardes2024vl}, while LLM-JEPA formulations replace or augment next-token prediction with latent-space semantic forecasting to induce state-level abstractions in language models \citep{goel2024llmjepa}. The key insight is that predicting in latent space, rather than reconstructing in input space, forces the encoder to capture abstract dynamics that reconstruction objectives can ignore. A pixel-prediction model can succeed by memorizing textures; a latent-prediction model must understand temporal continuity.

We introduce \textbf{SMB-Structure} to bring this insight to clinical foundation models. Our approach combines two primary objectives: supervised fine-tuning (SFT) reconstructs future patient states in token space, grounding the representation in clinical semantics, while the JEPA objective predicts those same futures in latent space from the current patient embedding alone. The critical difference between the SFT and JEPA objectives is when information about future states become available. In standard autoregressive training, the model can defer encoding trajectory dynamics until decoding time, relying on the decoder to translate the current representation into future tokens. With JEPA, the encoder must predict the future embedding before observing the future state, forcing dynamics into the representation itself. We validate this approach across two cohorts totaling over 40,000 patients, evaluating not at single timepoints but along disease trajectories. The results demonstrate that SMB-Structure captures dynamics not recoverable by autoregressive baselines, achieving competitive performance on tasks where patient heterogeneity makes trajectory-level reasoning essential.

\begin{figure*}[!ht]
\begin{center}
\includegraphics[width=0.75\textwidth]{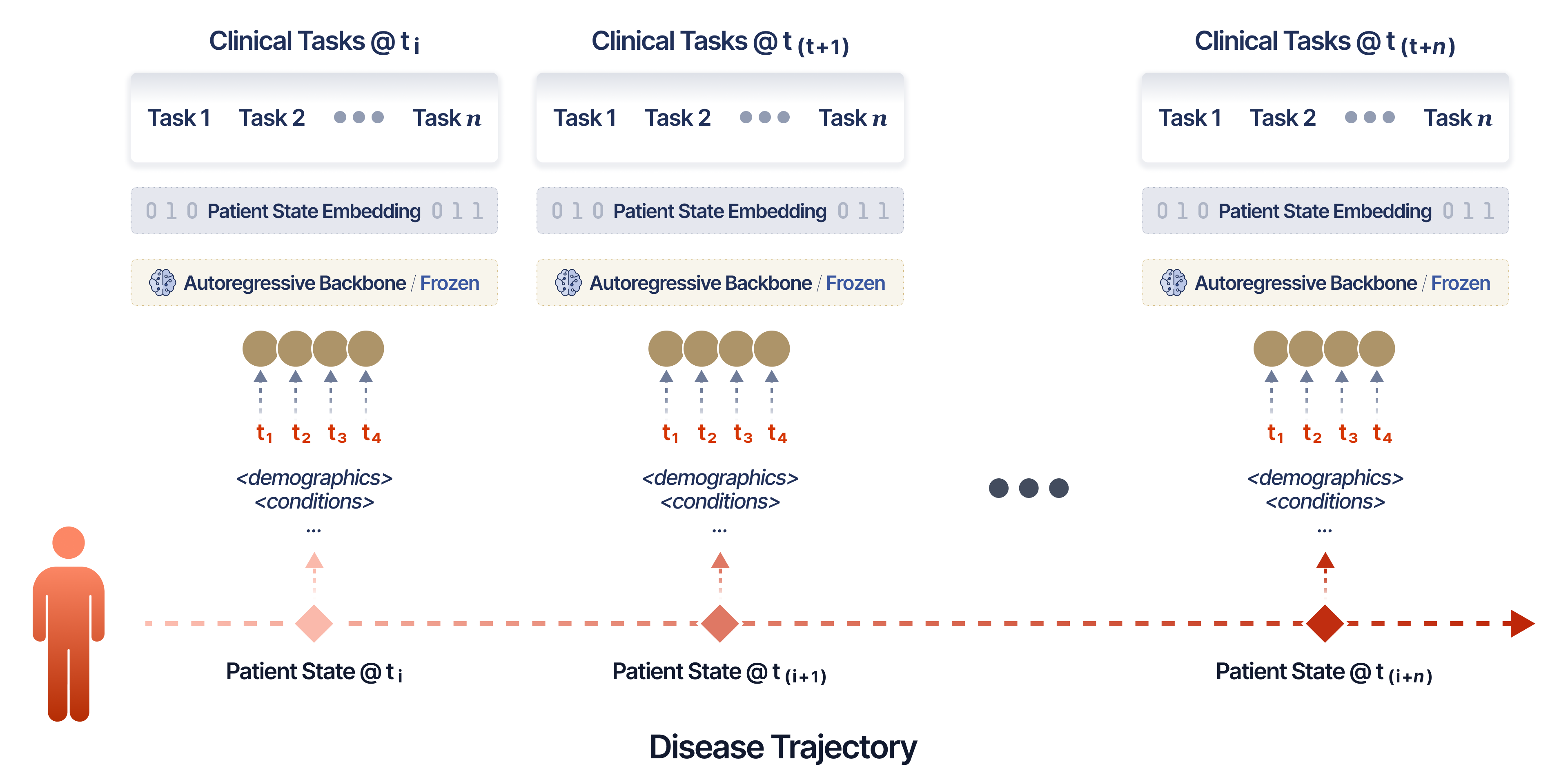}
\end{center}
\caption{Evaluation Framework of Foundation Models for Time-to-Event EHR.}
\label{fig:eval}  
\end{figure*}

%==============================================================================
\section{Related Work}
%==============================================================================

\subsection{Clinical EHR Foundation Models}

The application of foundation models to electronic health records has progressed along two complementary axes: models that process clinical text, and those that operate over structured medical codes. Text-based approaches adapt pretrained language models to clinical corpora, yielding strong performance on clinical report-level tasks such as named entity recognition, relation extraction, and document classification \citep{alsentzer2019publicly, peng2019transfer}. Large language models have since demonstrated emergent clinical knowledge: Med-PaLM achieves near-physician performance on medical licensing examinations through instruction tuning \citep{singhal2023large}, while multimodal variants integrate radiology images and pathology slides with clinical text \citep{chen2024advancing, li2024llavamed}. For structured EHR data, CLMBR introduced autoregressive pretraining over medical code sequences, demonstrating that representations from clinical language models transfer effectively to downstream prediction tasks \citep{steinberg2021language}. Recent work has explored next-event prediction as an enhanced objective for temporal reasoning \citep{chen2025building}, instruction tuning for conversational EHR assistants \citep{xu2024instruction}, and integration of multi-modal biomedical data for precision oncology \citep{chen2025patient, chen2023boosting}. Despite this progress, existing approaches share a common limitation: they optimize for reconstruction, predicting the next token, code, or event, rather than explicitly encoding how patient states evolve under interventions. Our work addresses this gap by grounding a Joint-Embedding Predictive Architecture with supervised fine-tuning, adapting latent-space prediction from world models to clinical trajectories.

\subsection{Joint-Embedding Predictive Architectures}

Joint-Embedding Predictive Architectures (JEPA) represent a distinct approach to self-supervised learning that predicts masked content in representation space rather than input space \citep{lecun2022path}. Unlike masked autoencoders that reconstruct pixels or tokens, JEPA models predict the embeddings of masked regions, abstracting away unpredictable surface details while preserving semantic structure, motivated by the insight that reconstruction objectives expend capacity modeling high-entropy, low-relevance variation rather than the abstract dynamics governing how inputs evolve \citep{lecun2022path,bengio2013representation}. This principle was first demonstrated in I-JEPA, which showed that predicting features rather than pixels yields representations with superior semantic transfer \citep{assran2023self}, and was extended to temporal domains by V-JEPA, where latent-space prediction captured motion dynamics, object permanence, and physical plausibility not encoded by pixel-reconstruction objectives \citep{bardes2024revisiting}. V-JEPA-2 further incorporated action-conditioned latent prediction, enabling zero-shot robotic planning by learning to simulate how visual states evolve under motor commands \citep{assran2025vjepa2}. More recently, JEPA-style objectives have been generalized to multimodal and language domains: VL-JEPA demonstrates that predicting aligned vision–language embeddings enables grounded cross-modal reasoning without explicit generative supervision \citep{bardes2024vl}, while emerging LLM-JEPA formulations replace or augment next-token prediction with latent-space semantic forecasting to induce state-level abstractions and temporal coherence in language models \citep{lecun2022path,goel2024llmjepa}. Collectively, these works position JEPA as a class of world models in which the predictor functions as an internal simulator, forecasting future latent states from current state and context \citep{ha2018world}. Unlike vision domains where JEPA operates over spatially coherent patches, EHR data presents heterogeneous, asynchronous observations, requiring us to ground JEPA with supervised fine-tuning to establish clinical semantics before learning trajectory dynamics. SMB-Structure thus combines the simulation-oriented objective of JEPA with the grounding of autoregressive language modeling, adapting latent-space dynamics to disease trajectories.

\section{Method}

\subsection{Problem Formulation}

We formalize a patient's clinical trajectory as a sequence of states $\mathbf{S} = \{s_1, s_2, \ldots, s_T\}$, where each state $s_t$ is encoded as a sequence of medical tokens spanning demographics, conditions, measurements, procedures, medications, and clinical notes. Given context $x = [x_1, \ldots, x_n]$ and future state $y = [y_1, \ldots, y_m]$, we seek an encoder $f_\theta$ whose representation $z = f_\theta(x)$ captures the dynamics governing $x \rightarrow y$.

Standard autoregressive training optimizes $p(y_t | y_{<t}, x)$, but this permits the encoder to defer trajectory reasoning to decode time; the decoder can rely on already-generated tokens $y_{<t}$ rather than encoding the full trajectory in $z$. We address this by requiring the encoder to predict future \emph{embeddings} before observing future tokens:
\begin{equation}
\hat{h}_i = g_\phi(f_\theta(\text{Mask}(x \oplus y, \mathcal{M}))) \approx f_{\bar{\theta}}(x \oplus y)_i \quad \forall i \in \mathcal{M}
\end{equation}
where $g_\phi$ is a predictor network, $f_{\bar{\theta}}$ is a momentum encoder providing stable targets, and $\mathcal{M}$ denotes masked positions in the target sequence.

\subsection{Model Architecture}

SMB-Structure extends a pretrained LLM backbone with three components: domain-specific clinical tokens, a bottleneck predictor for latent-space prediction, and a momentum encoder for stable targets (Figure~\ref{fig:arch}).

\paragraph{Clinical Tokenization.} Patient records are serialized using delimiter tokens that demarcate clinical categories (Table~\ref{tab:tokens}). These tokens are added to the vocabulary and their embeddings learned during fine-tuning, enabling the model to leverage structural information in EHR data. Similar structural tokenization has been adopted in prior longitudinal EHR language models to make record boundaries and heterogeneous fields explicit \citep{rajamohan2025gptehr, redekop2025zeroshot, kanchinadam2024mediclaimgpt}. Our section tags follow the same principle, but expose clinical field boundaries (demographics, diagnoses, labs, medications, notes) directly to the backbone.

\begin{table}[!ht]
\centering
\small
\begin{tabular}{ll}
\toprule
\textbf{Token} & \textbf{Content} \\
\midrule
\texttt{<demographics>} & Age, sex, ethnicity \\
\texttt{<conditions>} & ICD codes, diagnoses \\
\texttt{<measurements>} & Lab values, vitals \\
\texttt{<observations>} & Clinical observations \\
\texttt{<procedures>} & CPT codes, interventions \\
\texttt{<drugs>} & Medications, dosages \\
\texttt{<notes>} & Free-text clinical notes \\
\texttt{<death>} & Mortality events \\
\bottomrule'
\end{tabular}
\caption{Domain-specific delimiter tokens added to the vocabulary. Each token has a corresponding closing tag.}
\label{tab:tokens}
\end{table}

\paragraph{Predictor.} The predictor refines encoder representations to predict masked embeddings. We use a bottleneck architecture that projects to a lower dimension before processing, serving as an information constraint that encourages abstract predictive features rather than surface-level pattern matching. Masked positions are replaced with a learnable token $\mathbf{m} \in \mathbb{R}^d$, then:
\begin{align}
\mathbf{H}_{\text{bottleneck}} &= \tilde{\mathbf{H}} W_{\text{down}} \\
\hat{\mathbf{H}} &= \text{LayerNorm}\!\left(
\text{Transformer}(\mathbf{H}_{\text{bottleneck}})\, W_{\text{up}}
\right)
\end{align}

where $W_{\text{down}} \in \mathbb{R}^{d \times d_b}$ and $W_{\text{up}} \in \mathbb{R}^{d_b \times d}$, and $d_b$ is the bottleneck dimension. The mask token is initialized from $\mathcal{N}(0, 0.02)$.

\paragraph{Momentum Encoder.} Following JEPA \citep{assran2023self}, we maintain an EMA copy of the encoder updated as $\bar{\theta} \leftarrow \tau \bar{\theta} + (1 - \tau) \theta$ after each gradient step. The momentum encoder processes \emph{unmasked} inputs to provide target embeddings. This asymmetry is critical: without it, the encoder could collapse to trivial solutions where all representations are identical. The high momentum ($\tau = 0.996$) ensures targets change slowly, providing a stable learning signal.

\subsection{Training Objectives}

\subsubsection{Supervised Fine-Tuning (SFT)}

The SFT objective grounds the model in clinical semantics through next-token prediction on unmasked sequences:
\begin{equation}
\mathcal{L}_{\text{SFT}} = -\frac{1}{m}\sum_{t=1}^{m} \log p_\theta(y_t | y_{<t}, x)
\end{equation}
This ensures the encoder learns clinically meaningful representations and prevents the JEPA objective from driving toward degenerate solutions that are predictive in embedding space but semantically meaningless.

\subsubsection{JEPA Objective}
The JEPA objective forces trajectory dynamics into the representation by requiring prediction in latent space, rather than token space, before observing future states. We mask a fraction $r_m = 0.5$ of target tokens based on ablations (Table~\ref{tab:ablations}), applying masking exclusively to the continuation (future state) while preserving the full context.

\begin{table*}[!h]
\centering
\caption{Evaluation on MSK cohort (oncology): AUC-ROC by cancer indication for different learning objectives. We compare SFT-only (baseline) vs Hybrid (SFT+JEPA) vs Curriculum (SFT then JEPA); M denotes training on MSK and I denotes adding INSPECT.}
\label{tab:auc-by-indication-msk}
\small
\setlength{\tabcolsep}{3pt}
\renewcommand{\arraystretch}{1.15}
\scalebox{1.0}{%
\begin{tabular}{lccccccccc}
\toprule
% First Row: Multirow for "Learning Objectives" + Multicolumn for "Cancer Types"
\multirow{2}{*}{\textbf{Methods}} & 
\multicolumn{9}{c}{\textbf{Cancer Types}} \\
% Draw the line only across columns 2-10
\cmidrule(lr){2-10}
% Second Row: Empty first cell (placeholder for multirow) + Specific Cancers
 & \textbf{Bladder} & \textbf{Kidney} & \textbf{Lung} & \textbf{Ovarian} & \textbf{Pancreas} & \textbf{Prostate} & \textbf{Sarcoma} & \textbf{Upper-GI} & \textbf{Uterus} \\
\midrule
\multicolumn{10}{c}{\texttt{Baselines}} \\
\midrule
Logistic Regression & \perf{0.676}{0.076} & \perf{0.561}{0.125} & \perf{0.672}{0.109} & \perf{0.670}{0.163} & \perf{0.679}{0.066} & \perf{0.579}{0.158} & \perf{0.652}{0.123} & \perf{0.630}{0.023} & \perf{0.653}{0.105} \\
XGBoost & \perf{0.701}{0.144} & \perf{0.641}{0.098} & \perf{0.671}{0.110} & \perf{0.735}{0.102} & \perf{0.695}{0.046} & \perf{0.659}{0.094} & \perf{0.731}{0.108} & \perf{0.643}{0.043} & \perf{0.689}{0.105} \\
Random Forest & \perf{0.707}{0.141} & \perf{0.582}{0.131} & \perf{0.664}{0.106} & \perf{0.745}{0.133} & \perf{0.696}{0.048} & \perf{0.637}{0.107} & \perf{0.735}{0.106} & \perf{0.628}{0.043} & \perf{0.681}{0.109} \\
\midrule
\multicolumn{10}{c}{\texttt{LLaMA3.1 8B}} \\
\midrule
SFT~(M) & \perf{0.751}{0.021} & \perf{0.665}{0.060} & \perf{0.803}{0.174} & \perf{0.744}{0.049} & \perf{0.656}{0.128} & \perf{0.752}{0.009} & \perf{0.729}{0.004} & \perf{0.678}{0.019} & \perf{0.730}{0.029} \\
SFT~(M+I) & \perf{0.759}{0.021} & \perf{0.667}{0.059} & \perf{0.805}{0.161} & \perf{0.745}{0.058} & \perf{0.660}{0.100} & \perf{0.737}{0.007} & \perf{0.732}{0.001} & \perf{0.693}{0.053} & \perf{0.734}{0.018} \\
Hybrid~(M) & \perf{0.763}{0.019} & \textbf{\perf{0.691}{0.059}} & \perf{0.805}{0.173} & \perf{0.746}{0.055} & \perf{0.667}{0.112} & \perf{0.748}{0.013} & \perf{0.737}{0.011} & \perf{0.667}{0.033} & \perf{0.735}{0.023} \\
Hybrid~(M+I) & \ul{\perf{0.764}{0.015}} & \perf{0.674}{0.071} & \textbf{\perf{0.819}{0.161}} & \ul{\perf{0.757}{0.051}} & \perf{0.695}{0.099} & \textbf{\perf{0.749}{0.014}} & \perf{0.742}{0.003} & \textbf{\perf{0.693}{0.046}} & \textbf{\perf{0.744}{0.027}} \\
Curr.~(M) & \perf{\textbf{0.769}}{\textbf{0.019}} & \perf{0.675}{0.067} & \ul{\perf{0.818}{0.163}} & \textbf{\perf{0.758}{0.056}} & \textbf{\perf{0.702}{0.091}} & \perf{0.746}{0.018} & \textbf{\perf{0.743}{0.001}} & \perf{0.681}{0.068} & \ul{\perf{0.743}{0.032}} \\
Curr.~(M+I) & \perf{0.763}{0.017} & \ul{\perf{0.683}{0.074}} & \perf{0.816}{0.164} & \perf{0.755}{0.050} & \ul{\perf{0.695}{0.097}} & \ul{\perf{0.748}{0.012}} & \ul{\perf{0.742}{0.001}} & \ul{\perf{0.691}{0.044}} & \perf{0.739}{0.024} \\
\midrule
\multicolumn{10}{c}{\texttt{Qwen3 8B}} \\
\midrule
SFT~(M) & \perf{0.771}{0.017} & \perf{0.682}{0.069} & \perf{0.816}{0.164} & \perf{0.756}{0.048} & \perf{0.691}{0.101} & \ul{\perf{0.759}{0.010}} & \perf{0.744}{0.003} & \perf{0.686}{0.050} & \perf{0.744}{0.025} \\
SFT~(M+I) & \perf{0.770}{0.015} & \perf{0.683}{0.065} & \perf{0.815}{0.158} & \perf{0.756}{0.045} & \perf{0.692}{0.097} & \perf{0.754}{0.009} & \perf{0.745}{0.002} & \perf{0.688}{0.047} & \perf{0.745}{0.022} \\
Hybrid~(M) & \perf{0.752}{0.014} & \perf{0.687}{0.053} & \perf{0.801}{0.176} & \perf{0.744}{0.054} & \perf{0.671}{0.102} & \perf{0.756}{0.010} & \perf{0.717}{0.003} & \perf{0.666}{0.032} & \perf{0.723}{0.020} \\
Hybrid~(M+I) & \textbf{\perf{0.778}{0.008}} & \perf{0.686}{0.076} & \textbf{\perf{0.818}{0.164}} & \textbf{\perf{0.766}{0.044}} & \textbf{\perf{0.698}{0.096}} & \textbf{\perf{0.760}{0.034}} & \textbf{\perf{0.751}{0.009}} & \textbf{\perf{0.698}{0.055}} & \textbf{\perf{0.750}{0.035} }\\
Curr.~(M) & \perf{0.759}{0.011} & \textbf{\perf{0.692}{0.060}} & \perf{0.803}{0.174} & \perf{0.743}{0.053} & \perf{0.669}{0.116} & \perf{0.754}{0.004} & \perf{0.726}{0.009} & \perf{0.672}{0.032} & \perf{0.722}{0.017} \\
Curr.~(M+I) & \ul{\perf{0.773}{0.011}} & \ul{\perf{0.689}{0.068}} & \ul{\perf{0.817}{0.164}} & \textbf{\perf{0.766}{0.053}} & \ul{\perf{0.694}{0.102}} & \perf{0.748}{0.032} & \ul{\perf{0.749}{0.002}} & \ul{\perf{0.694}{0.068}} & \ul{\perf{0.748}{0.034}} \\
\bottomrule
\end{tabular}%
}
\label{tab:msk-indication}
\end{table*}

Unlike masked language modeling, SMB-Structure is never rewarded for reconstructing the masked tokens or their vocabulary distributions. Instead, the prediction target is the latent embedding produced by a slowly evolving momentum encoder on the unmasked sequence, which removes incentives to model surface-form statistics and encourages state-level abstractions that support forecasting. This asymmetric online/target design is crucial to avoid degenerate fixed points in which representations become trivially predictable.

The online encoder processes the masked sequence:
\begin{equation}
\mathbf{H}_{\text{online}} = f_\theta(x_1, \ldots, x_n, \tilde{y}_1, \ldots, \tilde{y}_m)
\end{equation}
where $\tilde{y}_i = \mathbf{m}$ for $i \in \mathcal{M}$. The momentum encoder processes the complete sequence to produce targets $\mathbf{H}_{\text{target}} = f_{\bar{\theta}}(x \oplus y)$. The JEPA loss minimizes MSE at masked positions:
\begin{equation}
\mathcal{L}_{\text{JEPA}} = \frac{1}{|\mathcal{M}|} \sum_{i \in \mathcal{M}} \| \hat{h}_i - \bar{h}_i \|_2^2
\end{equation}

The key distinction from SFT is temporal: the encoder must predict future embeddings \emph{without access to the masked tokens}, forcing it to encode trajectory dynamics in the context representation rather than deferring to decoding time.

\subsubsection{Joint Optimization}

We combine both objectives:
\begin{equation}
\mathcal{L} = \lambda_{\text{SFT}} \mathcal{L}_{\text{SFT}} + \lambda_{\text{JEPA}} \mathcal{L}_{\text{JEPA}}
\end{equation}
with $\lambda_{\text{SFT}} = 1.0$ and $\lambda_{\text{JEPA}} = 1.0$ based on ablations (Table~\ref{tab:ablations}).

Each training step involves two forward passes through the online encoder: Pass 1 computes SFT loss on the unmasked sequence; Pass 2 computes the JEPA loss on the masked sequence, with the momentum encoder (gradients disabled)  providing targets. Gradients from both passes are accumulated before a single optimizer step, after which the momentum encoder is updated via EMA. Refer to Appendix~\ref{tab:impl} for implementation details.

\section{Experiments}
We evaluate SMB-Structure on its ability to capture general-purpose trajectory dynamics across diverse clinical outcomes. Training is entirely label-free, relying only on self-supervised objectives over patient trajectories (Figure~\ref{fig:arch}). Crucially, all downstream tasks (68 from MSK, 7 from INSPECT) are predicted from a frozen embedding $S(t)$ using only linear probes. This \textit{single-embedding}, \textit{multi-task} protocol ensures that performance gains reflect the intrinsic quality of the learned representations rather than task-specific fine-tuning or catastrophic forgetting of the latent world model.

\begin{table}[]
\caption{Evaluation on MSK cohort: AUC-ROC across task categories (disease progression, toxicity/adverse events, treatment outcomes) comparing learning objectives (SFT baseline, Hybrid SFT+JEPA, Curriculum SFT\,$\rightarrow$\,JEPA) and training sets (M=MSK, I=INSPECT).}
\centering
\small
\setlength{\tabcolsep}{3pt}
\renewcommand{\arraystretch}{1.15}
\scalebox{1.0}{%
\begin{tabular}{lccc}
\toprule
\textbf{Methods} & 
\begin{tabular}{@{}c@{}}\textbf{Disease} \\ \textbf{Progression}\end{tabular} & 
\begin{tabular}{@{}c@{}}\textbf{Toxicity \&} \\ \textbf{Adverse Events}\end{tabular} & 
\textbf{Mortality} \\
\midrule
\multicolumn{4}{c}{\texttt{LLaMA3.1 8B}} \\
\midrule
SFT~(M) & \perf{0.727}{0.054} & \perf{0.734}{0.231} & \perf{0.740}{0.019} \\
SFT~(M+I) & \perf{0.725}{0.053} & \perf{0.722}{0.226} & \perf{0.743}{0.016} \\
Hybrid~(M) & \perf{0.719}{0.047} & \perf{0.723}{0.244} & \perf{0.735}{0.021} \\
Hybrid~(M+I) & \perf{0.727}{0.050} & \ul{\perf{0.741}{0.226}} & \textbf{\perf{0.746}{0.017}} \\
Curr.~(M) & \perf{0.716}{0.046} & \textbf{\perf{0.742}{0.252}} & \perf{0.731}{0.019} \\
Curr.~(M+I) & \textbf{\perf{0.731}{0.048}} & \perf{0.738}{0.228} & \textbf{\perf{0.746}{0.016}} \\
\midrule
\multicolumn{4}{c}{\texttt{Qwen3 8B}} \\
\midrule
SFT~(M) & \perf{0.728}{0.050} & \perf{0.732}{0.231} & \perf{0.748}{0.016} \\
Hybrid~(M) & \perf{0.718}{0.046} & \perf{0.727}{0.239} & \perf{0.725}{0.019} \\
Hybrid~(M+I) & \ul{\perf{0.728}{0.051}} & \textbf{\perf{0.743}{0.224}} & \textbf{\perf{0.761}{0.017}} \\
Curr.~(M) & \perf{0.719}{0.044} & \perf{0.724}{0.245} & \perf{0.731}{0.020} \\
Curr.~(M+I) & \textbf{\perf{0.730}{0.049}} & \ul{\perf{0.736}{0.233}} & \ul{\perf{0.755}{0.015}} \\
\bottomrule
\end{tabular}%
}
\vspace{-10pt}
\label{tab:msk-tasks}
\end{table}

\subsection{Datasets}
We utilize two large-scale longitudinal real-world cohorts that provide the temporal resolution necessary to model disease as a dynamical system.

\subsubsection{Memorial Sloan Kettering (MSK) Oncology}
To evaluate the model's capacity for high-dimensional trajectory modeling, we leverage a massive, real-world cohort from the MSK iHub Challenge\footnote{https://www.mskcc.org/commercialization/programs-accelerators/msk-innovation-hub/msk-ihub-challenge}. The dataset comprises 23,319 patients and over 323,000 patient-years across 9 major cancer indications (see Appendix~\ref{tab:msk-cohort-stats} for more details). With a median follow-up of 50 months and an average of 127 clinical events per patient, MSK provides a high-fidelity longitudinal record. The near-universal coverage of pathology (100\%) and deep genomic biomarkers (95.6\%) allows SMB-Structure to model complex interactions between systemic therapies and the evolving biological state of the tumor.

\subsubsection{INSPECT (Pulmonary Embolism)}
% 19,402 pulmonary embolism patients
% Tasks: acute trajectory prediction
We further validate on INSPECT~\citep{huang2023inspect}, a multimodal dataset specifically designed for pulmonary embolism (PE) diagnosis and prognosis. While the full dataset includes CTPA images and radiology reports, we utilize the longitudinal EHR component, which contains de-identified records for 19,402 patients and over 225 million medical events. INSPECT serves as a critical benchmark for acute and sub-acute prognosis, averaging 5,080 records per patient from initial diagnosis through long-term outcomes.

\subsection{Evaluation Protocol}
Clinical decision-making occurs along a shifting trajectory of risk rather than at a single classification point. To mirror this reality, we adopt a Point-in-Time framework (see Figure~\ref{fig:eval}) that evaluates the model at critical clinical junctions, or "Decision Nodes," as outlined in Table~\ref{tab:trigger-events}.

The evaluation slices each patient’s history into multiple snapshots at various decision nodes. For each node, an indexing date ($t=0$) is established. The model consumes the longitudinal narrative up to $t=0$ and generates a single fused patient state $S(t)$. Crucially, all data occurring after $t=0$ is strictly masked. A linear probe, such as a Cox Proportional Hazards loss~\cite{katzman2018deepsurv} for survival tasks, is then tasked with predicting the future trajectory based solely on the latent information contained within $S(t)$.
To ensure rigorous validation, we enforce three specific safeguards: a patient-level split (85/15 ratio by Subject ID) to prevent subject overlap; a 24-hour temporal buffer between $t=0$ and the prediction window to eliminate intraday leakage; and an end-of-life filter excluding data within seven days of mortality events to remove administrative cues associated with terminal care.  We report the mean and standard deviation of AUC-ROC scores aggregated by clinical category or disease indication to summarize performance stability across the 68 distinct downstream  (see Appendix~\ref{tab:downstream-tasks}).

\begin{table}[]
\caption{\textbf{Indexing Date Triggers on MSK.} The generator scans patient event logs for these five specific clinical triggers to establish a \textit{decision node} ($t=0$) where future data is masked.}
\centering
\small
\setlength{\tabcolsep}{3pt}
\renewcommand{\arraystretch}{1.4}
\scalebox{0.8}{%
\begin{tabular}{ll}
\toprule
\textbf{Trigger Event} & \textbf{Description} \\
\midrule
Start of Therapy & The initiation of a new systemic regimen. \\
Confirmed Progression & Radiographic or clinical evidence of disease worsening. \\
Curative Surgery & Major interventions that reset recurrence risk. \\
Metastatic Diagnosis & The transition from localized to systemic disease. \\
Performance Decline & Significant drops in ECOG or KPS scores. \\
\bottomrule
\end{tabular}%
}
\vspace{-10pt}
\label{tab:trigger-events}
\end{table}

\begin{table*}[]
\centering
\caption{Evaluation on INSPECT cohort (PE): AUC-ROC by cancer indication for different learning objectives. We compare SFT-only (baseline) vs Hybrid (SFT+JEPA) vs Curriculum (SFT then JEPA); M denotes training on MSK and I denotes adding INSPECT.}
\label{tab:auc-by-indication-msk}
\small
\setlength{\tabcolsep}{3pt}
\renewcommand{\arraystretch}{1.2}
\scalebox{0.9}{%
\begin{tabular}{lccccccc}
\toprule
% Row 1: Group Headers
\multirow{2}{*}{\textbf{Methods}} & 
\multicolumn{3}{c}{\textbf{Mortality}} & 
\multicolumn{1}{c}{\textbf{PH}} & 
\multicolumn{3}{c}{\textbf{Readmission}} \\
% Rules: Separate the groups (leaves a gap between them)
\cmidrule(lr){2-4} \cmidrule(lr){5-5} \cmidrule(lr){6-8}
% Row 2: Specific Timepoints
 & 30 day & 180 day & 365 day & 365 day & 30 day & 180 day & 365 day \\
\midrule
\multicolumn{8}{c}{\texttt{LLaMA3.1 8B}} \\
\midrule
SFT~(M) & 0.790 & 0.793 & 0.800 & 0.645 & 0.678 & 0.672 & 0.674 \\
SFT~(M+I) & 0.792 & 0.794 & 0.802 & 0.629 & 0.677 & 0.672 & 0.674 \\
Hybrid~(M) & 0.743 & 0.733 & 0.741 & 0.603 & 0.628 & 0.621 & 0.623 \\
Hybrid~(M+I) & \ul{0.795} & \ul{0.796} & \ul{0.803} & \ul{0.634} & \ul{0.681} & \ul{0.676} & \ul{0.676} \\
Curr.~(M) & 0.736 & 0.722 & 0.732 & 0.605 & 0.620 & 0.617 & 0.619 \\
Curr.~(M+I) & \textbf{0.806} & \textbf{0.803} & \textbf{0.810} & \textbf{0.653} & \textbf{0.691} & \textbf{0.681} & \textbf{0.680} \\
\midrule
\multicolumn{8}{c}{\texttt{Qwen 8B}} \\
\midrule
SFT~(M) & 0.781 & 0.778 & 0.788 & \ul{0.625} & 0.655 & 0.659 & 0.663 \\
SFT~(M+I) & 0.782 & 0.779 & 0.788 & 0.623 & 0.656 & 0.659 & 0.663 \\
Hybrid~(M) & 0.722 & 0.701 & 0.708 & 0.589 & 0.596 & 0.596 & 0.604 \\
Hybrid~(M+I) & \ul{0.783} & \ul{0.781} & \ul{0.793} & 0.618 & \ul{0.669} & \ul{0.667} & \ul{0.669} \\
Curr.~(M) & 0.725 & 0.704 & 0.710 & 0.594 & 0.594 & 0.593 & 0.601 \\
Curr.~(M+I) & \textbf{0.790} & \textbf{0.786} & \textbf{0.797} & \textbf{0.625} & \textbf{0.679} & \textbf{0.671} & \textbf{0.671} \\
\bottomrule
\end{tabular}
}
\vspace{-10pt}
\label{tab:inspect}
\end{table*}

\subsection{Task Taxonomy}
Across the two cohorts, we evaluate a comprehensive suite of \textbf{68} distinct tasks designed to test whether the embedding encodes "momentum", the direction and velocity of a patient's health status.

The MSK cohort includes \textbf{61} tasks spanning three primary categories: Disease Progression (e.g., site-specific worsening), Toxicity and Adverse Events (e.g., Grade 3+ events), and Survival (e.g., all-cause mortality). The assessment of disease progression under treatment is particularly critical, as it requires the model to simulate how a specific tumor might react to a given therapeutic agent.

The INSPECT cohort features \textbf{7} tasks focused on acute prognosis across three categories: mortality (assessed at 30, 180, and 365-day intervals), readmission (similarly tiered), and chronic complications, specifically the development of Pulmonary Hypertension.

\subsection{Baselines \& Training Methods}
We compare SMB-Structure against a spectrum of established clinical learning paradigms. We begin with traditional discriminative models, including Logistic Regression, Random Forest, and XGBoost—trained on flattened, feature-engineered patient profiles. These baselines serve to quantify the predictive ceiling of non-sequential, tabular approaches. Moving to sequential modeling, we evaluate an SFT-only baseline, representing the standard autoregressive clinical LLM optimized solely for next-token prediction; this effectively tests the "Patient as a Document" hypothesis. Finally, to isolate the impact of our proposed learning dynamics, we examine two internal variants: \textbf{SMB-Structure Hybrid}, which optimizes reconstruction (SFT) and latent prediction (JEPA) objectives simultaneously, and \textbf{SMB-Structure Curriculum}, which employs a two-stage protocol to establish semantic grounding before introducing latent trajectory modeling.

\subsection{Results \& Discussions}
\subsubsection{Main Results}
Our experiments were designed to test the central hypothesis of this work: that clinical foundation models require explicit trajectory modeling (JEPA) to capture dynamics that are invisible to standard next-token prediction (SFT). The results across MSK and INSPECT (Tables \ref{tab:msk-indication},~\ref{tab:msk-tasks},~\ref{tab:inspect}, and Appendix Figures~\ref{fig:benchmark}, ~\ref{fig:lung_cancer}) validate this hypothesis, but with critical nuances regarding training stability and data diversity.

\textbf{Trajectory Diversity as a Regularizer.} The interaction between training data scale and objective function reveals a distinct advantage for the JEPA formulation. Adding the INSPECT cohort (M+I) yields negligible gains for the SFT baseline (e.g., MSK treatment outcomes remain static at $0.740\rightarrow0.743$). This indicates that the SFT model has likely saturated its ability to learn from distributional regularities~\citep{balestriero2025lejepa}; simply seeing more tokens does not improve its \textit{world model}.

In contrast, the JEPA-based objectives (Hybrid and Curriculum) exhibit a strong \textit{Trajectory Regularization} effect. The addition of INSPECT, which contains acute, rapid-evolution trajectories of Pulmonary Embolism, significantly boosts performance on the chronic, slower-moving oncology tasks in MSK (e.g., Hybrid LLaMA mortality prediction jumps from $0.735$ to $0.746$).

This suggests that while SFT learns \textit{dataset-specific} token statistics, the JEPA objective learns \textit{universal} dynamics of physiological change. Exposure to the rapid decay and recovery patterns in PE helps the model better encode the subtle progression signals in oncology. The latent dynamics model generalizes across diseases better than the token reconstruction model generalizes across vocabularies.

\begin{table}[htb]
\centering
\caption{Ablation studies on Qwen3-1.7B (MSK cohort). $\text{Pred}_{\text{D}}$: predictor depth; $\text{Pred}_{\text{W}}$: predictor width; $\lambda_{\text{J}}:\lambda_{\text{S}}$: JEPA-to-SFT loss ratio; $r_m$: masking ratio. \textbf{Bold}: best within each ablation.}
\label{tab:ablations}
\small
\setlength{\tabcolsep}{6pt}
\renewcommand{\arraystretch}{1.1}
\scalebox{1.0}{%
\begin{tabular}{@{}ccccc@{}}
\toprule
$\text{Pred}_{\text{D}}$ & $\text{Pred}_{\text{W}}$ & $\lambda_{\text{J}}:\lambda_{\text{S}}$ & $r_m$ & \textbf{MSK AUC} \\
\midrule
\multicolumn{5}{l}{\textit{SFT Only (Baseline)}} \\
\midrule
-- & -- & -- & -- & 0.716 \\
\midrule
\multicolumn{5}{l}{\textit{(a) Predictor Architecture}} \\
\midrule
1 & $0.5h$ & 1:1 & 1.0 & 0.718 \\
1 & $1.0h$ & 1:1 & 1.0 & 0.720 \\
1 & $2.0h$ & 1:1 & 1.0 & 0.721 \\
2 & $0.5h$ & 1:1 & 1.0 & 0.719 \\
2 & $1.0h$ & 1:1 & 1.0 & \textbf{0.724} \\
2 & $2.0h$ & 1:1 & 1.0 & 0.722 \\
4 & $0.5h$ & 1:1 & 1.0 & 0.722 \\
4 & $1.0h$ & 1:1 & 1.0 & 0.722 \\
4 & $2.0h$ & 1:1 & 1.0 & 0.723 \\
\midrule
\multicolumn{5}{l}{\textit{(b) Loss Weighting}} \\
\midrule
2 & $1.0h$ & 1:2 & 1.0 & 0.723 \\
2 & $1.0h$ & 1:1 & 1.0 & \textbf{0.725} \\
2 & $1.0h$ & 2:1 & 1.0 & 0.722 \\
2 & $1.0h$ & 2.5:1 & 1.0 & 0.722 \\
\midrule
\multicolumn{5}{l}{\textit{(c) Masking Ratio}} \\
\midrule
2 & $1.0h$ & 1:1 & 0.25 & 0.725 \\
2 & $1.0h$ & 1:1 & 0.50 & \textbf{0.728} \\
2 & $1.0h$ & 1:1 & 0.75 & 0.724 \\
2 & $1.0h$ & 1:1 & 1.00 & 0.725 \\
\bottomrule
\end{tabular}
}
\label{tab:ablations}
\vspace{-15pt}
\end{table}

% \vspace{-14pt}
\textbf{Capturing "Clinical Momentum": Long-Horizon vs. Short-Horizon.} The clearest evidence that SMB-Structure captures \textit{momentum}, which is the velocity of disease progression rather than just static risk factors lies in the time-horizon sensitivity on the INSPECT cohort (Table \ref{tab:inspect}).

On short-term tasks like 30-day readmission, the gap between Curriculum~(M+I) and SFT~(M+I) is noticeable but modest ($0.691$ vs. $0.677$, a $+2.0\%$ relative improvement). However, on 365-day mortality, requiring long-range extrapolation of the patient's current trajectory, the gap widens significantly ($0.810$ vs. $0.802$).

This differential performance validates the architectural motivation outlined in the Introduction. SFT models excel at identifying explicit risk markers present in the immediate context (e.g., a "hospice referral" token), making them competitive on short-term tasks. However, they struggle to model the latent rate of change required for long-term forecasting. By forcing the encoder to predict future embeddings without access to future tokens, SMB-Structure explicitly encodes this velocity, allowing it to sustain predictive accuracy over longer horizons where static cues have faded.

\textbf{The Necessity of Semantic Grounding: Curriculum vs. Hybrid.} A striking finding in Table \ref{tab:msk-tasks} is the failure of joint optimization (Hybrid) to consistently outperform the SFT baseline when trained on a single cohort. For example, Hybrid~(M) lags behind SFT~(M) on MSK disease progression ($0.719$ vs. $0.727$ AUC-ROC with LLaMA-3.1).

This performance degradation suggests a fundamental \textit{objective interference} phenomenon. The SFT objective pushes representations to preserve high-frequency local details necessary for token reconstruction, while the JEPA objective pushes representations to abstract away noise and encode slowly-evolving state dynamics. When optimized simultaneously from scratch, these gradients appear to conflict, preventing the encoder from mastering either.

The success of the curriculum strategy ($0.731$ AUC on disease progression) resolves this conflict. By treating SFT as a \textit{semantic bootstrapping} phase, the model first establishes a stable clinical vocabulary and grounding. The subsequent JEPA phase then refines these representations, shifting the encoder’s focus from what the patient is (static description) to where the patient is going (dynamic prediction). This confirms that latent world modeling is most effective when applied to a representation space that is already semantically structured.

\subsubsection{Ablation Studies}We examine the impact of architectural and optimization choices on the Qwen3-1.7B backbone (Table~\ref{tab:ablations}), identifying the specific constraints required to encode clinical dynamics.

\textbf{Predictor Complexity.} Results indicate that a 2-layer MLP with width matching the LLM hidden dimension ($1.0h$) acts as an optimal transition operator ($0.724$ AUC). Shallower predictors ($0.718$-$0.721$) lack the non-linearity to model complex physiological transitions, while deeper networks ($0.722$) yield diminishing returns, suggesting the primary bottleneck lies in encoder representation quality rather than the depth of the transition function itself.

\textbf{Objective Balancing.} An equal weighting between SFT and JEPA objectives ($\lambda_{\text{J}}:\lambda_{\text{S}} = 1:1$) proves most robust ($0.725$ AUC). Skewing towards SFT causes the model to over-index on static token statistics, while skewing towards JEPA allows representations to drift from the grounded clinical vocabulary, decoupling the latent space from the downstream linear probes.

\textbf{Masking Ratio.} We identify a clear optimum at $r_m = 0.50$ (+1.2\% over baseline). Lower ratios ($r_m=0.25$) render the task trivial by allowing leakage from local correlations, whereas excessive masking ($r_m \ge 0.75$) removes the context necessary for logical deduction, making the trajectory structurally unpredictable. The findings confirm that performance gains stem from a carefully tuned information bottleneck rather than increased parameter counts.

\section{Conclusion}
We presented \textbf{SMB-Structure}, a training paradigm for longitudinal EHR that integrates supervised fine-tuning with a Joint-Embedding Predictive Architecture, bridging the gap between clinical documentation reconstruction and trajectory simulation. By enforcing latent-space forecasting before observing future states, our approach compels the encoder to capture dynamics often ignored by standard autoregressive objectives. Validated across 40,000 patients, SMB-Structure achieves competitive performance on long-horizon prediction tasks, confirming that separating semantic grounding from dynamical modeling yields superior representations for heterogeneous patient histories.

While currently limited by the computational overhead of dual forward passes and an evaluation scope restricted to linear probing, this work establishes a critical foundation for next-generation clinical AI. Future iterations will extend this framework to intervention-conditioned world models to enable counterfactual reasoning and treatment optimization. Ultimately, we demonstrate that effective clinical reasoning requires models that both understand what a patient record says and simulate where the patient's disease state is going.

\section*{Impact Statement}

This work develops foundation models for clinical prediction from electronic health records. Potential benefits include earlier identification of disease progression and treatment response, which could improve patient outcomes and resource allocation. However, clinical prediction models carry risks: biases in training data may perpetuate disparities across patient populations, and overreliance on model predictions could reduce clinician autonomy. Our models were trained on data from two institutions (MSK and Stanford) with specific patient demographics; generalization to other populations requires careful validation. We do not recommend deployment without prospective evaluation, fairness audits across demographic subgroups, and integration with clinical workflows that preserve physician oversight. All data used in this study were de-identified and accessed under institutional review board approval.

\nocite{langley00}

\bibliography{main}
\bibliographystyle{icml2026}

%%%%%%%%%%%%%%%%%%%%%%%%%%%%%%%%%%%%%%%%%%%%%%%%%%%%%%%%%%%%%%%%%%%%%%%%%%%%%%%
%%%%%%%%%%%%%%%%%%%%%%%%%%%%%%%%%%%%%%%%%%%%%%%%%%%%%%%%%%%%%%%%%%%%%%%%%%%%%%%
% APPENDIX
%%%%%%%%%%%%%%%%%%%%%%%%%%%%%%%%%%%%%%%%%%%%%%%%%%%%%%%%%%%%%%%%%%%%%%%%%%%%%%%
%%%%%%%%%%%%%%%%%%%%%%%%%%%%%%%%%%%%%%%%%%%%%%%%%%%%%%%%%%%%%%%%%%%%%%%%%%%%%%%
\newpage
\appendix
\onecolumn
\section{Appendix}

\subsection{Implementation Details}

Table~\ref{tab:impl} summarizes our configuration. We fine-tuned with LoRA applied to all linear layers in the backbone, adding $\sim$167M trainable parameters. The predictor contributes $\sim$67M additional parameters. For curriculum training (SMB-Structure Curriculum), we first train with SFT only to establish clinical language understanding, then introduce the JEPA objective to learn predictive dynamics.

\begin{table}[htb]
\centering
\small
\setlength{\tabcolsep}{5pt}
\renewcommand{\arraystretch}{1.4}
\scalebox{1.0}{%
\begin{tabular}{ll}
\toprule
\textbf{Component} & \textbf{Configuration} \\
\midrule
\multicolumn{2}{l}{\textit{Model}} \\
Model Family & LLaMA-3 / Qwen-3 \\
LoRA & $r=64$, $\alpha=128$, dropout $0.1$ \\
Predictor & 2 layers, $d_b=h_{llm}$, 8 heads, GELU \\
\midrule
\multicolumn{2}{l}{\textit{JEPA}} \\
Mask ratio $r_m$ & 0.5 \\
Momentum $\tau$ & 0.996 \\
Loss weights $\lambda_{\text{SFT}}:\lambda_{\text{JEPA}}$ & $1.0:1.0$ \\
\midrule
\multicolumn{2}{l}{\textit{Training}} \\
Hardware & 8$\times$ H200 (DeepSpeed ZeRO-1) \\
Batch size & 13 per GPU \\
Schedule & Cosine decay, 3\% warmup \\
Optimizer & AdamW ($\beta_1=0.9$, $\beta_2=0.95$) \\
Weight decay & 0.1 \\
Max sequence length & 3,300 tokens \\
Precision & BF16 \\
\bottomrule
\end{tabular}
}
\caption{Implementation details for SMB-Structure.}
\label{tab:impl}
\end{table}

\begin{table*}[]
\centering 
\small 
\caption{Comprehensive Statistics for the MSK Oncology Cohort ($N=23,319$)} \label{tab:msk-cohort-stats} 
\setlength{\tabcolsep}{5pt}
\renewcommand{\arraystretch}{1.4}
\scalebox{1.0}{%
\begin{tabular}{lllc} \toprule \textbf{Category} & \textbf{Metric / Feature} & \textbf{Value} & \textbf{Coverage} \\ \midrule \multirow{4}{*}{\textbf{General Overview}} & Total Clinical Records & $3,003,408$ & -- \\ & Total Clinical Volume & $323,833$ Patient-Years & -- \\ & Median Follow-up & $50.7$ Months & -- \\ & Data Density & $127$ Events/Patient (Avg) & -- \\ \midrule \multirow{5}{*}{\textbf{Top Indications}} & Lung Cancer & $570,710$ Records & -- \\ & Ovarian Cancer & $539,715$ Records & -- \\ & Pancreas Cancer & $387,658$ Records & -- \\ & Prostate Cancer & $326,761$ Records & -- \\ & Uterus Cancer & $309,650$ Records & -- \\ \midrule \multirow{6}{*}{\textbf{OMOP Domain Coverage}} & Person / Condition / Observation & -- & $100.0\%$ \\ & Note / NLP / Procedure & -- & $100.0\%$ \\ & Measurement (Labs) & -- & $99.6\%$ \\ & Drug Exposure & -- & $85.8\%$ \\ & Death (Mortality) & -- & $42.9\%$ \\ \midrule \multirow{10}{*}{\textbf{Oncology Modalities}} & Pathology & $16.1$ Records/Patient & $100.0\%$ \\ & Outcomes (Survival Data) & $2.1$ Records/Patient & $100.0\%$ \\ & Clinical Assessment (ECOG/KPS) & $38.9$ Records/Patient & $99.9\%$ \\ & Imaging & $17.9$ Records/Patient & $99.1\%$ \\ & Biomarkers (Genomics) & $3.3$ Records/Patient & $95.6\%$ \\ & Systemic Therapy & $18.2$ Records/Patient & $94.8\%$ \\ & Local Therapy (Surg/Rad) & $4.8$ Records/Patient & $86.5\%$ \\ & Staging (TNM) & $10.2$ Records/Patient & $78.0\%$ \\ & Toxicity (Adverse Events) & $13.4$ Records/Patient & $33.2\%$ \\ & Complications & $22.9$ Records/Patient & $24.5\%$ \\ \bottomrule \end{tabular}
}
\end{table*}

\begin{table}[]
\caption{\textbf{Downstream Clinical Evaluation Tasks.} To validate the robustness of the Point-in-Time embeddings, we evaluate performance across five distinct clinical dimensions using a linear probe.}
\centering
\small
\setlength{\tabcolsep}{5pt}
\renewcommand{\arraystretch}{1.4}
\scalebox{1.0}{%
\begin{tabular}{ll p{4.5cm} p{5.5cm}}
\toprule
\textbf{Task Dimension} & \textbf{Window} & \textbf{Prediction Definition} & \textbf{Clinical Significance} \\
\midrule
\textbf{Disease} & 180 Days & Predict site-specific worsening & Tests if the embedding captures the \\
\textbf{Progression} & & (e.g., Lung, Liver, Bone). & spatial and temporal momentum of the tumor. \\
\midrule
\textbf{Toxicity \&} & 90 Days & Predict Grade 3+ toxicity & Tests the patient's physiological \\
\textbf{Adverse Events} & & or treatment discontinuation. & reserve and tolerance, distinct from tumor dynamics. \\
\midrule
\textbf{Treatment} & 365 Days & Predict initiation of the next & A proxy for ``Time to Treatment \\
\textbf{Durability} & & line of therapy (e.g., 1st to & Failure,'' capturing the interaction \\
& & 2nd line transition). & between disease and intervention. \\
\midrule
\textbf{Treatment} & 180 Days & Predict therapeutic success & Verifies capacity to simulate \\
\textbf{Response} & & based on RECIST outcomes & direct causal effects rather than \\
& & (e.g., ``Remission''). & natural disease history. \\
\midrule
\textbf{Overall} & 365 Days & Predict all-cause mortality. & The ultimate composite metric \\
\textbf{Survival} & & & of disease severity. \\
\bottomrule
\end{tabular}%
}
\label{tab:downstream-tasks}
\end{table}

\begin{figure}[htb]
    \centering
    \includegraphics[width=\linewidth]{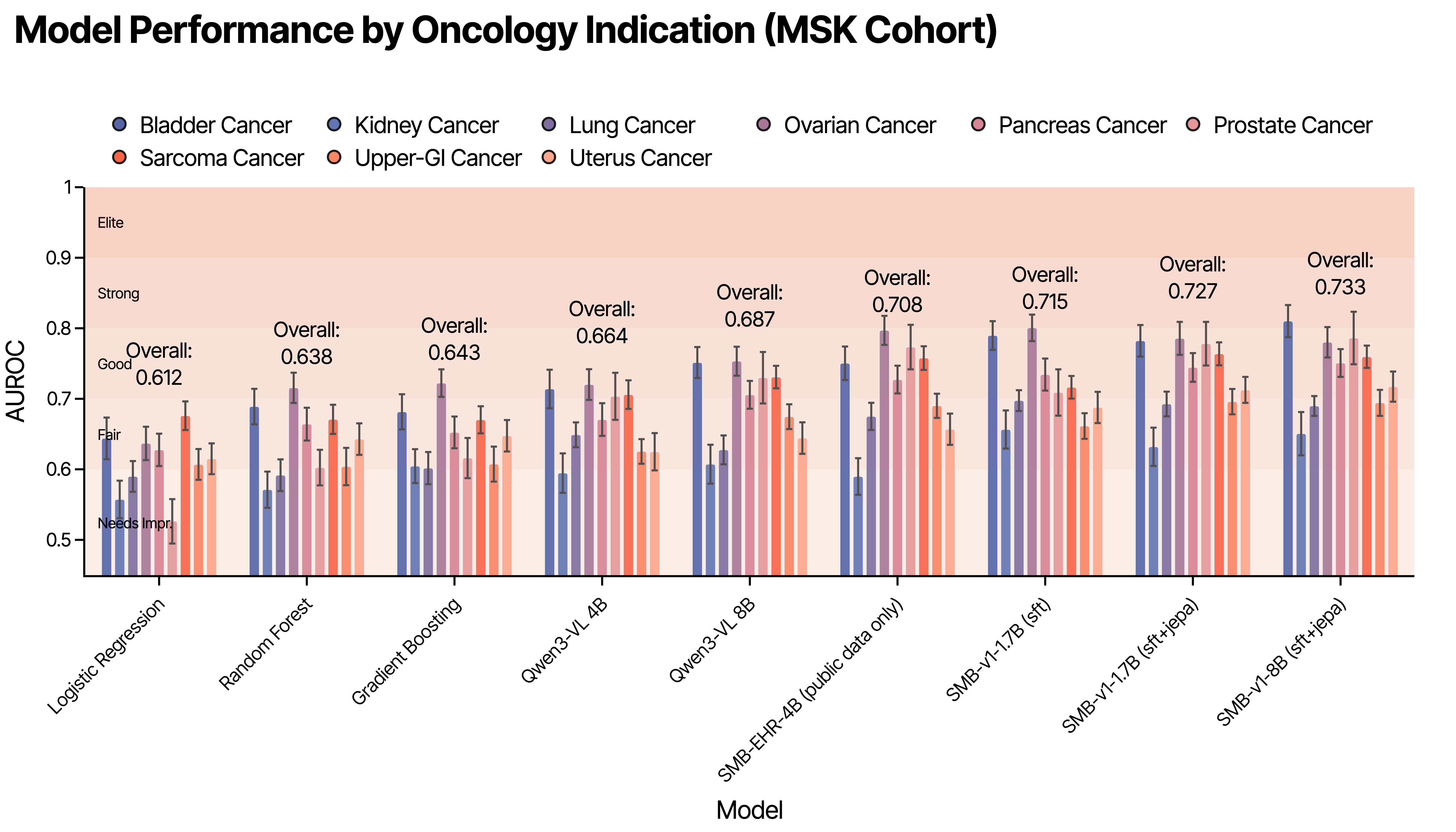}
    \caption{Model performance by oncology indication on MSK cohort.}
    \label{fig:benchmark}
\end{figure}

\begin{figure}[htb]
    \centering
    \includegraphics[width=\linewidth]{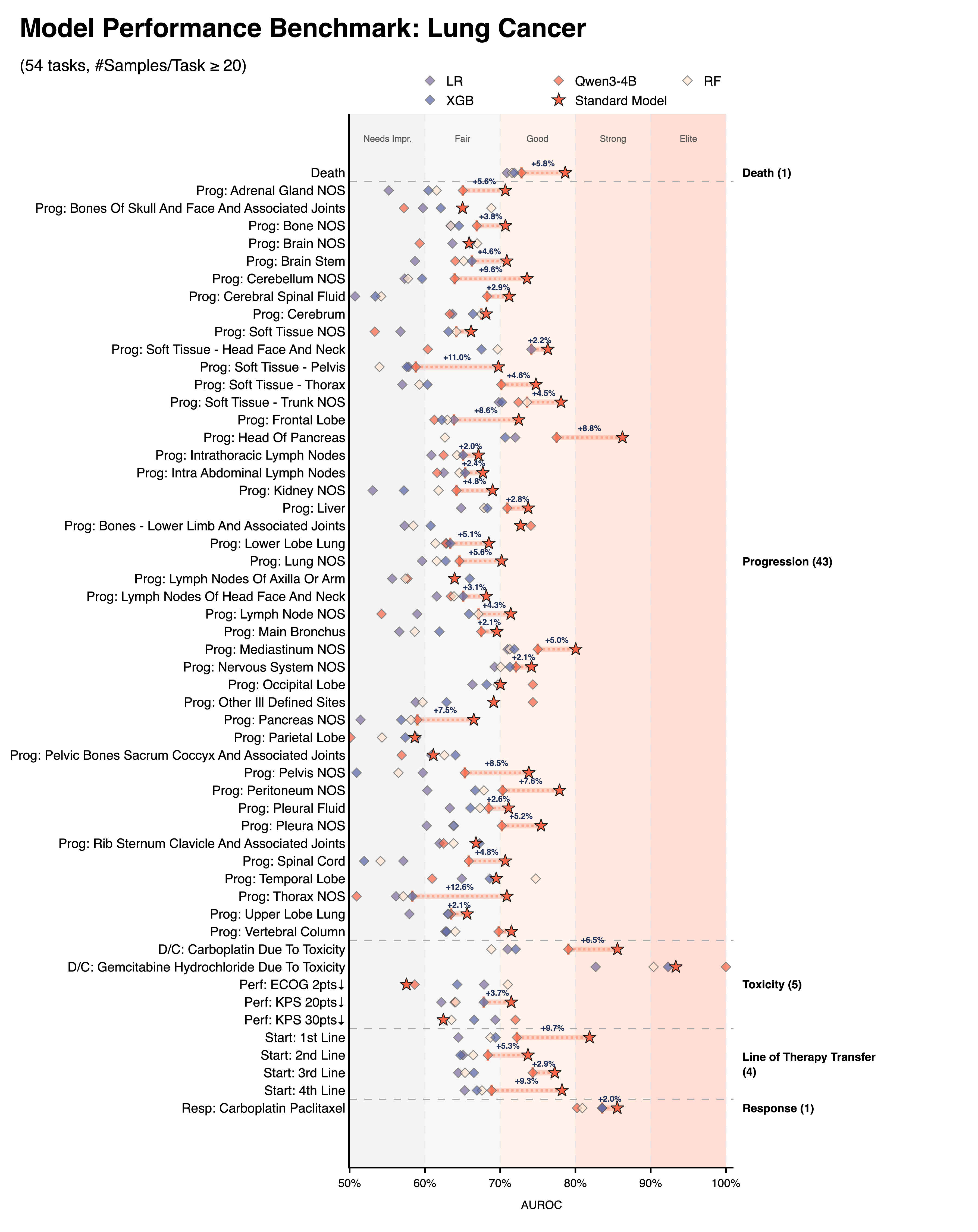}
    \caption{Model performance benchmark on Lung Cancer.}
    \label{fig:lung_cancer}
\end{figure}

%%%%%%%%%%%%%%%%%%%%%%%%%%%%%%%%%%%%%%%%%%%%%%%%%%%%%%%%%%%%%%%%%%%%%%%%%%%%%%%
%%%%%%%%%%%%%%%%%%%%%%%%%%%%%%%%%%%%%%%%%%%%%%%%%%%%%%%%%%%%%%%%%%%%%%%%%%%%%%%

\end{document}

% This document was modified from the file originally made available by
% Pat Langley and Andrea Danyluk for ICML-2K. This version was created
% by Iain Murray in 2018, and modified by Alexandre Bouchard in
% 2019 and 2021 and by Csaba Szepesvari, Gang Niu and Sivan Sabato in 2022.
% Modified again in 2023 and 2024 by Sivan Sabato and Jonathan Scarlett.
% Previous contributors include Dan Roy, Lise Getoor and Tobias
% Scheffer, which was slightly modified from the 2010 version by
% Thorsten Joachims & Johannes Fuernkranz, slightly modified from the
% 2009 version by Kiri Wagstaff and Sam Roweis's 2008 version, which is
% slightly modified from Prasad Tadepalli's 2007 version which is a
% lightly changed version of the previous year's version by Andrew
% Moore, which was in turn edited from those of Kristian Kersting and
% Codrina Lauth. Alex Smola contributed to the algorithmic style files.